\documentclass[runningheads]{llncs}
\usepackage{graphicx}

\usepackage{amsmath,amsfonts,bm}

\def\eqref#1{equation~\ref{#1}}

\def\1{\bm{1}}

\DeclareMathAlphabet{\mathsfit}{\encodingdefault}{\sfdefault}{m}{sl}
\SetMathAlphabet{\mathsfit}{bold}{\encodingdefault}{\sfdefault}{bx}{n}

\usepackage{graphicx}
\usepackage{hyperref}
\usepackage{url}
\usepackage[utf8]{inputenc} 
\usepackage[T1]{fontenc}    
\usepackage{booktabs}       
\usepackage{amsfonts}       
\usepackage{nicefrac}       
\usepackage{microtype}      
\usepackage{amsmath}
\usepackage{xspace}
\usepackage{tabularx}
\usepackage{multicol}
\usepackage{multirow}

\begin{document}
\title{Disentangling Latent Factors of Variational Auto-Encoder with Whitening}
%
%
\author{Sangchul Hahn \and
Heeyoul Choi}
\authorrunning{S. Hahn and H. Choi.}
%
\institute{Handong Global University, Pohang, South Korea \\
\email{\{s.hahn, hchoi\}@handong.edu}\\
}
\maketitle              
\begin{abstract}
After deep generative models were successfully applied to image generation tasks, learning disentangled latent variables of data has become a crucial part of deep generative model research. Many models have been proposed to learn an interpretable and factorized representation of latent variable by modifying their objective function or model architecture. To disentangle the latent variable, some models show lower quality of reconstructed images and others increase the model complexity which is hard to train. In this paper, we propose a simple disentangling method based on a traditional whitening process. The proposed method is applied to the latent variables of variational auto-encoder (VAE), although it can be applied to any generative models with latent variables. In experiment, we apply the proposed method to simple VAE models and experiment results confirm that our method finds more interpretable factors from the latent space while keeping the reconstruction error the same as the conventional VAE's error. 

\end{abstract}

\section{Introduction}
\label{sec:intro}
Since variational auto-encoder (VAE) \cite{vae-kingma} and generative adversarial network (GAN) \cite{gan-ian} were proposed in deep learning, various deep generative models have been introduced and they have shown remarkable results in image generation tasks \cite{info-gan,beta-vae,factor-vae}. Once generative models become successful, disentangling the latent variable of the models has been another major research point. Since the latent variables can play a role as conceptual factors, disentangled latent variables make the models understand our world more conceptually \cite{representation-learning,info-gan,beta-vae}. 

As variants of the VAE structure, $\beta$-VAE \cite{beta-vae} and Factor-VAE \cite{factor-vae} change the original objective function to make their latent variable more factorized than the original VAE model. Note that the original VAE model itself enforces the latent variable to be factorized based on the isotropic Gaussian prior. As one of the GAN networks, Info-GAN \cite{info-gan} adds a feature latent code to the original input latent variable to learn independent factors of input data. 

Although these models show remarkable results in learning an explainable and factorized latent variable, they have drawbacks. For example, $\beta$-VAE achieves a better disentangling result \cite{factor-vae} at the cost of lower reconstruction quality compared to the original VAE. Factor-VAE overcomes the drawback of $\beta$-VAE by introducing a new disentangling method to VAE, but it needs an additional network structure (discriminator). Also, Info-GAN provides good reconstruction quality and disentangling result. However, because of the GAN structure, it has an unstable training issue. Even with W-GAN \cite{w-gan}, which provides more stability to training GAN, there are still some issues in hyper-parameter tuning.

In this paper, we introduce a new disentangling method which is simple and easy to apply to any deep generative models with latent variables. 
The original generative models are good to capture important factors behind the input data, but one problem is that each dimension of the latent variables is correlated to others so that it is hard to figure the meaning of individual dimensions. Therefore, if we can make the latent variables uncorrelated to each other, we could obtain more disentangled or explainable latent variables. Whitening with principal component analysis (PCA) is one of the most frequently used methods that converts a set of correlated variables into a set of linearly uncorrelated variables. In this point of view, we propose to apply PCA whitening to the latent variable of the original VAE. 

To verify our method, we compare it to other methods on several image datasets, qualitatively and quantitatively. The qualitative analysis of disentanglement is based on just encoding the input image and generating images while traversing each dimension's value of the latent variables. If the generated images are changing by only one factor of the images when we change one dimension of the latent variables, it means the latent variables are well disentangled or factorized \cite{representation-learning}. 
However, despite the growing interests in research of disentanglement, there is a lack of standard quantitative evaluation metric, although a few papers suggested evaluation metrics recently \cite{beta-vae,eastwood2018a,factor-vae}. In this paper, we use the evaluation metric proposed in \cite{factor-vae} for quantitative verification. We apply our proposed method to the original VAE model and compare to three other models (VAE, $\beta$-VAE, and Factor-VAE) on three datasets (MNIST, CelebA, 2D Shapes) \cite{mnist_lecun,celeba,dsprites17}.

The paper is organized as follows. We introduce background knowledge including deep generative models and PCA whitening in section 2. In section 3, we review related works like $\beta$-VAE, Factor-VAE, Info-GAN. Then, we describe our method to disentangle the latent variables of the models in section 4. The datasets and models that we used in our experiments and the results are presented in section 5. Finally, we conclude this paper with a summary of our work and future works in section 6.

\section{Related Work}
\label{sec:related}

\subsection{Deep Generative Models}
\label{ssec:dgm}
Deep generative models are based on deep neural networks and aim to learn a true data distribution from training data in the unsupervised learning manner. If the model can learn the true data distribution, it is possible to generate new data samples from the learned distribution with some variations. However, sometimes it is not possible to learn the true data distribution. Therefore, deep generative models train neural networks to approximate the true data distribution, which leads to model distribution. 

In recent deep learning, most deep generative models are variations of VAE or GAN. VAE is an auto-encoder with a constraint on the latent space which is forced to be isotropic Gaussian by minimizing the Kullback-Leibler (KL) divergence between the Gaussian prior and the model distribution. Since the latent space generates samples for the decoder, the reparameterization trick is applied to make the gradient information flow through the latent space. After training the model, the latent space keeps most information to reconstruct input data, as well as it becomes isotropic Gaussian as much as possible. In other words, VAE cannot have perfectly disentangled latent variable even with the isotropic Gaussian prior. Contrary to VAE, in the conventional GAN models, there is no constraint on the latent space. Thus, in both VAE and GAN, the dimensions of the latent space might be entangled with other dimensions. 

\subsection{Whitening with PCA}
\label{ssec:whitening}
Whitening with PCA is a preprocessing step to make data linearly uncorrelated. PCA whitening is composed of two steps. First, it applies PCA to the data samples to transform the correlated data distribution to uncorrelated one. Second, it normalizes each dimension with the square root of the corresponding eigenvalue to make each dimension have the unit variance.

\subsection{Disentangling Models}
\label{ssec:disentangling_models}
Provided that $z$ is the latent variable of the model and $x$ is the input data, VAE models are trained to maximize the following objective function.
\begin{equation}
\mathcal{L}_{VAE} = \mathbb{E}_{q(z|x)}[\log p(x|z)] - KL(q(z|x)||p(z)),
\label{eq:vae}
\end{equation}
where $KL(q||p)$ means the KL divergence between $q$ and $p$, and $q(z|x)$, $p(x|z)$ and $p(z)$ are the encoder, decoder, and the prior distributions, respectively. The encoder and decoder are implemented with deep neural networks, and the prior distribution is isotropic Gaussian with unit variance. See \cite{vae-kingma} for the details.  

As variations of VAE, several disentangling models have been proposed \cite{beta-vae,factor-vae}. $\beta$-VAE changes the original objective function of VAE with a new parameter $\beta$ as in Eq. \ref{eq:beta_vae}. 
\begin{equation}
\mathcal{L}_{\beta\_VAE} = \mathbb{E}_{q(z|x)}[\log p(x|z)] - \beta(KL(q(z|x)||p(z))).
\label{eq:beta_vae}
\end{equation} 
When $\beta = 1$, it is exactly the same as the original objective function of VAE as in Eq. \ref{eq:vae}. However, with $\beta > 1$, it constrains the expression power of the latent variable $z$, and makes the distribution of $z$ to be more similar to the isotropic Gaussian distribution. That is, KL divergence becomes more important with higher $\beta$ values.  
Also, if $\beta$ becomes larger, the latent variables can be more disentangled by resembling the isotropic Gaussian. That is, the KL divergence term in the objective function of $\beta$-VAE encourages conditional independence (or uncorrelatedness) in $q(z|x)$ \cite{beta-vae}. However, there is a trade-off between reconstruction error and disentanglement \cite{factor-vae}. In other words, the quality of reconstruction is damaged with larger $\beta$ values, with which the latent variable can be more disentangled. 

To overcome the side effect of $\beta$-VAE, Factor-VAE proposes another disentangling method based on the VAE structure \cite{factor-vae}. In addition to the objective function of original VAE, Factor-VAE adds another KL divergence term to regularize the latent variables to be factorized as in Eq. \ref{eq:factor_vae}. 
\begin{multline}
\mathcal{L}_{F\_VAE} = \mathbb{E}_{q(z|x)}[\log p(x|z)] - KL(q(z|x)||p(z)) - \gamma KL(q(z)||\bar{q}(z)),
\label{eq:factor_vae}
\end{multline}
where 
\begin{eqnarray}
q(z) &=&\mathbb{E}_{p(x)}[q(z|x)], \\
\bar{q}&=&\prod_{j=1}^{d}q(z_j),    
\label{eq:q_z}
\end{eqnarray}
and $d$ is the dimension of the latent variables. In other words, the second KL divergence term is added to force $q(z)$ to be as independent as possible.
Since $\mathbb{E}_{p(x)}[q(z|x)]$ is intractable in practice, it is approximated by sampling $N$ input cases as follows. 
\begin{equation}
\mathbb{E}_{p(x)}[q(z|x)] \approx \frac{1}{N}\sum_{i=1}^{N}q(z|x^{(i)}).
\end{equation}
Note that the first two terms in Eq. \ref{eq:factor_vae} are exactly the same as the original VAE objective function, Eq. \ref{eq:vae}.

\subsection{Metrics for disentanglement}
\label{ssec:metric}
Despite the growing interests about disentangling models, there is no standard evaluation metric and lack of labeled data for evaluation. Therefore, the previously proposed disentangling models verify their disentangling quality based on qualitative analysis. Most commonly used analysis is based on latent variable traversal. If only one factor of generated images is changing while changing the value of one dimension in the latent variable, then the latent variable that the model learned is considered to be well disentangled. This qualitative analysis is easy to understand and intuitive, but we still need quantitative analysis methods to compare the disentangling ability of various models. 

Recently several evaluation metrics have been proposed for disentanglement with labeled data for that metrics \cite{beta-vae,eastwood2018a,factor-vae}. 
One of these quantitative evaluation metrics proposed by \cite{factor-vae} is summarized in Table. \ref{table:metric}. In experiments, we use this metric to compare models quantitatively. For the details of the metric, see Appendix B in \cite{factor-vae}.

\begin{table}[ht]
\caption{Algorithm: Disentangling metric.}
\label{table:metric}
  \centering
  \begin{tabular}{|l|}
\hline
1. Select $L$ images $(x_1, x_2, \dots , x_L)$ from $D_{f_k}$. \\
~~~$D_{f_k}$ is a set of sample images with a fixed value for the $k$-th generating factor \\
~~~and random value for the other factors. \\
2. Encode $(x_1, x_2, \dots, x_L)$ to make latent variables 
 $(z_1, z_2, \dots, z_L)$, where $z_i \in \mathbb{R}^d$ \\
3. Rescale latent variables with empirical standard deviation $s \in \mathbb{R}^d$.\\
4. Calculate empirical variance in each dimension of the rescaled latent variable.\\
5. Find a dimension, $d^*$, which has the minimum variance \\
~~~$d^*=\operatorname*{arg\,min}_d \mbox{var}(z_d)$. \\
6. Add $(d^*,k)$ into the training set.\\
7. Repeat 1 to 6 for all generating factors to make $M$ training votes.\\
~~~($M$ is the number of training votes) \\
8. Making a majority vote classifier with training votes and calculate accuracy \\
~~~of the majority vote classifier for the disentangling score. \\
\hline
\end{tabular}
\end{table}

\section{Whitening the Latent Variable}
\label{sec:whitening_latent}
We propose a new disentangling method based on whitening the latent variables with PCA, which leads to {\em Whitening VAE} (WVAE). 
We apply PCA to the original latent variables of the trained model, then we rescale the dimensions of the projected variable with the square root of corresponding eigenvalues. This is the process of PCA based whitening method. With the rescaled eigenspace, every dimension in the eigenspace has a unit variance. The reason why we rescale the latent space is to make the scale of control panel to be similar while we control the latent space for the latent variable traversal.

After training a generative model (VAE) with training data $\mathbf{X}$, the generative model encodes $\mathbf{X}$ to $\mathbf{Z}$ in the latent space. Then the whitening process is applied to $\mathbf{Z}$, which is summarized in Table \ref{table:algorithm}. Because our proposed method applies the whitening method to the latent space of the already trained model, it does not change the objective function of the original VAE. That is, the objective function of our model (Eq. \ref{eq:objective_fn}) is the same as one of the original VAE, and the reconstruction error also does not change. This means that our proposed method does not sacrifice the reconstruction quality while achieving more disentangled latent variables.
\begin{equation}
\mathcal{L}_{WVAE} = \mathbb{E}_{q(z|x)}[\log p(x|z)] - KL(q(z|x)||p(z)).
\label{eq:objective_fn}
\end{equation}

\begin{table}[ht]
\caption{Algorithm: Whitening the latent variable.}
\label{table:algorithm}
  \centering
  \begin{tabular}{|l|}
\hline
Given a trained VAE model. \\ 
1. Apply the encoder to $\mathbf{X}$ to obtain $\mathbf{Z}$. \\
2. Apply PCA to $\mathbf{Z}$ to find the eigenvalue $\Lambda$
and eigenvector $\mathbf{U}$ \\ 
 ~~~~~~ of the covariance matrix. \\
3. Project $\mathbf{Z}$ to the eigenspace by $\mathbf{Z}_{PCA} = \mathbf{U}^T \mathbf{Z}$.\\
4. Rescale the eigenspace by $\mathbf{Z}_{PCA\_W}=\Lambda^{1/2}\mathbf{Z}_{PCA}$ \\
\hline
\end{tabular}
\end{table}

\begin{figure}[ht!]
\centerline{\hbox{ 
\includegraphics[width=3.0in]{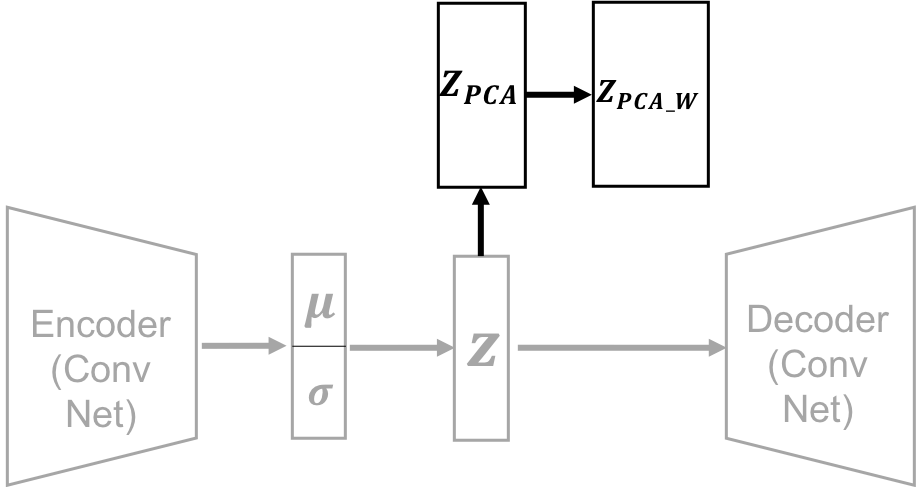}
}}
\centerline{\hbox{ (a) Model structure of WVAE}}
\vspace{0.3in}
\centerline{\hbox{ 
\includegraphics[width=3.0in]{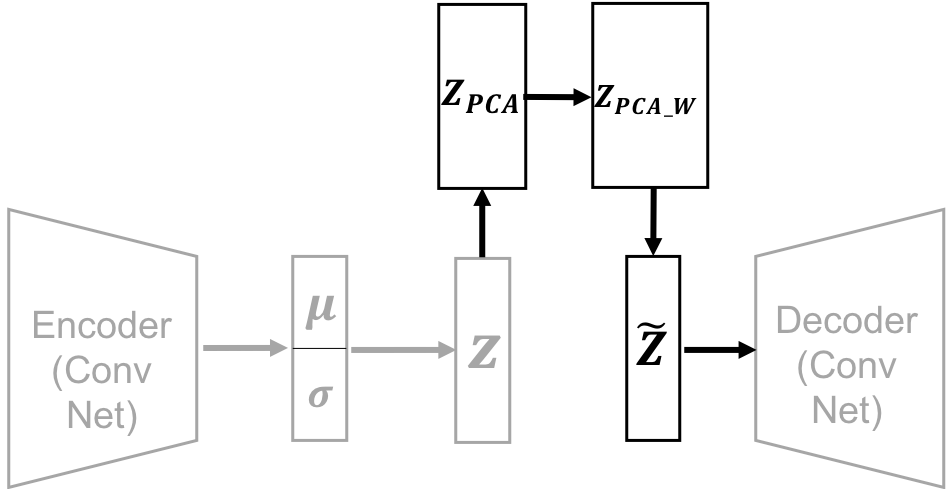}
}}
\centerline{\hbox{ 
(b) Process of latent variable traversals}}
\caption{The proposed model architecture. (a) Model structure of our proposed model with the process of making disentangled latent variable (WVAE) after training the VAE model. (b) process of latent variable traversals on WVAE. When doing latent variable traversals, the dimensions of $Z_{PCA\_W}$ are controled and recovered to the $\mathbf{Z}$ space to make changed latent variable $\tilde{Z}$. The VAE model part (in gray color) is fixed after training in advance. }
\label{fig:w_vae_structure}
\end{figure}

Fig. \ref{fig:w_vae_structure}(a) shows the model structure of our proposed method applied to VAE, and Fig. \ref{fig:w_vae_structure}(b) describes how latent variable traversal can be applied. To generate new samples, we control $\mathbf{Z}_{PCA\_W}$ in the rescaled eigenspace.

While $\beta$-VAE has a trade-off between disentangling and reconstruction quality as we described above, our proposed method does not sacrifice the reconstruction quality to disentangle the latent variable.  
Also, Factor-VAE needs sampling and approximation, because $\gamma KL(q(z)||\bar{q}(z))$ term is intractable in practice \cite{factor-vae}. Additionally, extra discriminator for the density-ratio trick is necessary to minimize the KL divergence term as in Eq. \ref{eq:factor_vae_discriminator} \cite{NguyenWJ10,Sugiyama2012}.
\begin{eqnarray}
TC(z)=KL(q(z)||\bar{q}(z))&=&\mathbb{E}_{q(z)}[\log \frac{q(z)}{\bar{q}(z)}]\nonumber\\
& \approx & \mathbb{E}_{q(z)}[\log \frac{D(z)}{1-D(z)}],
\label{eq:factor_vae_discriminator}
\end{eqnarray}
where $TC(z)$ is the total correlation \cite{Watanabe60} and $D$ is the discriminator. However, our proposed method does not need any extra network, sampling or approximation process.

\section{Experiments}
\label{sec:experiments}
\subsection{Data}
\label{ssec:data}
\begin{figure}[ht!]
\centerline{\hbox{ 
\includegraphics[width=3.5in]{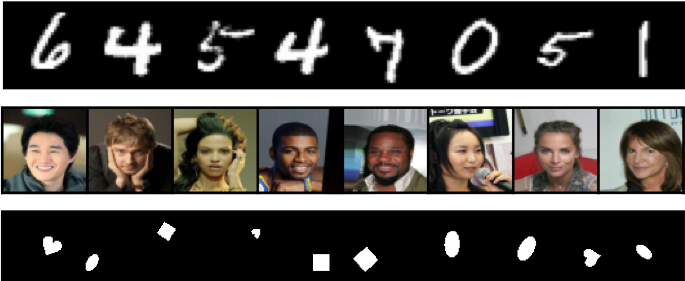}
}}
\caption{Image examples sampled from the three datasets. (Top) MNIST, (Middle) CelebA, and (Bottom) 2D Shapes.}
\label{fig:data_sample}
\end{figure}

We use three different image datasets for our experiments: MNIST, CelebA, and 2D Shapes. These datasets are most frequently used in many papers of deep generative models and disentanglement of latent variable. The first two datasets (MNIST and CelebA) have no label for generative factors, while the 2D shapes dataset has labels for generative factors. MNIST consists of 60K and 10K hand written images (28 $\times$ 28) for training and testing, respectively. CelebA (aligned and cropped version) has 202,599 RGB face images (64 $\times$ 64 $\times$ 3) of celebrities. 2D Shape has 737,280 images (64 $\times$ 64) that are generated with 6 generative factors (number of values): color(1), shape(3), scale(6), orientation(40), position X(32), and position Y(32). 2D Shape has label of generated factors as the value of each generative factors, so it can be used for quantitative analysis of disentangling performance. Fig. \ref{fig:data_sample} shows a few sample images of the datasets. We analyze qualitatively the methods on the three datasets, and analyze quantitatively on the 2D Shapes dataset. We compare our proposed method to three different methods: VAE, $\beta$-VAE, and Factor-VAE.

\subsection{Models}
\label{ssec:models}
To compare our proposed method to other models, we use the same VAE architecture of $\beta$-VAE, and Factor-VAE. The encoder consists of convolutional neural networks, and the decoder consists of deconvolutional neural networks. For Factor-VAE, we use fully connected layers for the discriminator as in \cite{factor-vae}. To apply the whitening process to the latent variable, we calculate eigenvalue and eigenvector with latent variable of the entire training data of each dataset. To disentangle the latent variable, our proposed model does not need any other architecture but need original VAE architecture only. We used RMSprop as an optimizer and set the learning rate to 0.001 at the beginning of the training process. We set the dimension of latent variable to 10 for all three datasets.

\subsection{Results}
\label{ssec:results}
\begin{figure}[ht!]
\centerline{\hbox{ 
\includegraphics[width=2.2in]{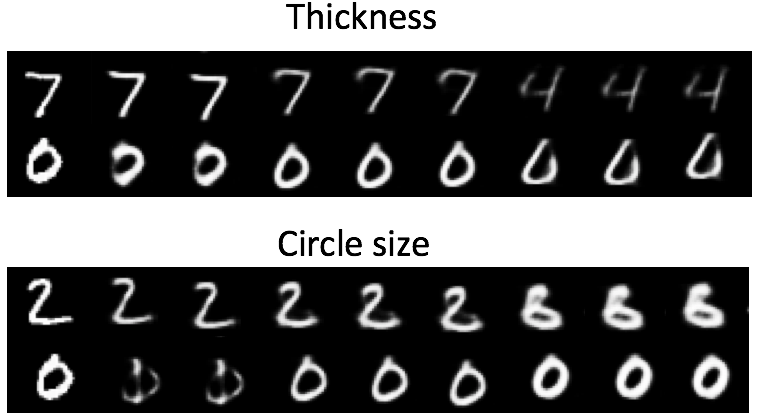}
\includegraphics[width=2.2in]{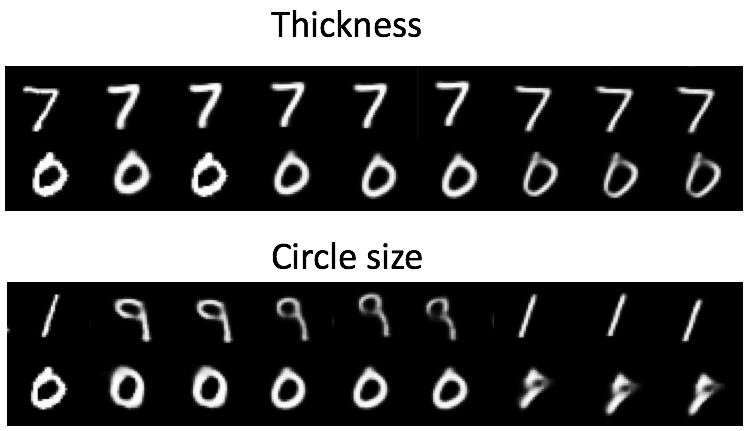}
}}
\centerline{\hbox{ 
(a) VAE  \hspace{1.7in} (b) WVAE  }}
\centerline{\hbox{ 
\includegraphics[width=2.2in]{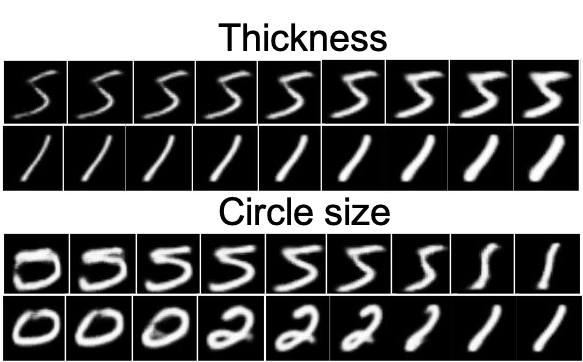}
\includegraphics[width=2.2in]{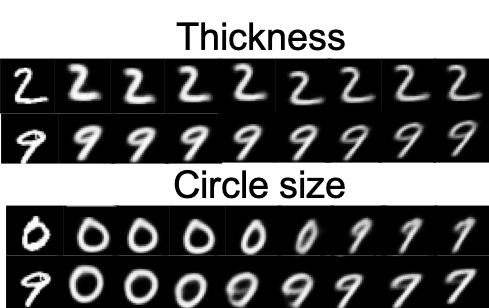}
}}
\centerline{\hbox{ 
(c) Factor-VAE  \hspace{1.5in} (d) $\beta$-VAE   }}
\caption{Qualitative analysis of conventional VAE, WVAE, Factor-VAE, and $\beta$-VAE on MNIST. The left-most column of each image is the ground truth image.}
\label{fig:qualitative_mnist}
\end{figure}
Fig. \ref{fig:qualitative_mnist} shows that the latent variable of our model is disentangled better than original VAE with MNIST. When we change the value of the dimension of a latent variable which corresponds to the factor of thickness, the generated image from original VAE changes with thickness and shape simultaneously, but the generated image from WVAE model changes with thickness only. As with the thickness, the image from original VAE changes with circle size and thickness when we change the circle factor but the image from WVAE changes with circle size only. This experiment results show that applying whitening method to latent variable can make latent variable more disentangled. Also, comparing to the Factor-VAE and $\beta$-VAE, it shows that our WVAE can disentangle the latent variable as good as Factor-VAE and $\beta$-VAE.

\begin{figure}[ht]
\centerline{\hbox{ 
\includegraphics[width=2.3in]{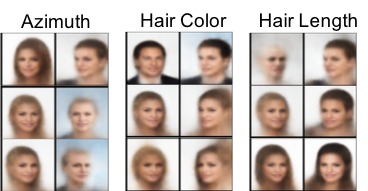}
\includegraphics[width=2.2in]{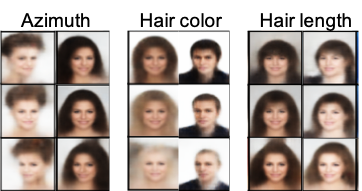}
}}
\centerline{\hbox{ 
(a) VAE \hspace{1.5in} (b) WVAE} \hspace{0.3in}}
\vspace{0.1in}
\centerline{\hbox{ 
\includegraphics[width=2.2in]{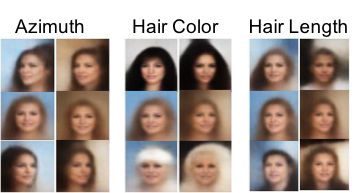}
\includegraphics[width=2.3in]{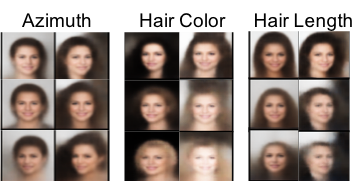}
}}
\centerline{\hbox{ 
  (c) Factor-VAE \hspace{1.5in} (d) $\beta$-VAE
}}
\caption{Qualitative analysis of VAE, WVAE, Factor-VAE, and $\beta$-VAE models on CelebA.}
\label{fig:qualitative_othermodels}
\end{figure}

Fig. \ref{fig:qualitative_othermodels} presents the results of latent variable traversals with four different models. 
The latent variable of the original VAE model is highly entangled. Factor-VAE ($\gamma=6.4)$ and $\beta$-VAE ($\beta=4$) show more distinct variations than original VAE. Comparing to the other models (VAE, $\beta$-VAE, and Factor-VAE), disentangling quality of our proposed method is as good as, if not better than, Factor-VAE and $\beta$-VAE, while it is much better than the original VAE. Note that our method is much simpler approach than the others.

\begin{figure}[ht]
\centerline{\hbox{ 
\includegraphics[width=1.0in,height=2.0in]{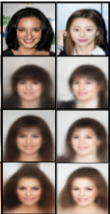}
\hspace{0.2in}
\includegraphics[width=1.0in,height=2.0in]{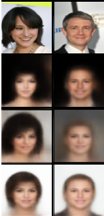}
\hspace{0.2in}
\includegraphics[width=1.0in,height=2.0in]{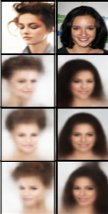}
}}
\centerline{\hbox{
(a)  \hspace{1.0in} (b)  \hspace{1.0in} (c) 
}}
\vspace{0.1in}
\centerline{\hbox{
\includegraphics[width=1.0in,height=2.0in]{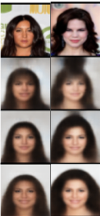}
\hspace{0.2in}
\includegraphics[width=1.0in,height=2.0in]{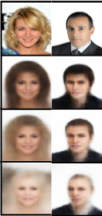}
}}
\centerline{\hbox{ 
(d) \hspace{1.0in} (e)}}
\caption{Qualitative analysis of our proposed method on the CelebA dataset. The top row of each image is the ground truth image. (a-e) show several results of latent variable traversals. The columns correspond to variations of hair length, background darkness, azimuth, smile, and hair color changing, respectively.}
\label{fig:qualitative_w_vae}
\end{figure}

Fig. \ref{fig:qualitative_w_vae} presents the results of latent variable traversals with our proposed method. Each image ((a)-(e)) correspond to variations of hair length, background darkness, azimuth, smile, and hair color changing, respectively, and the variations are distinct. It shows that our whitened latent variables have linearly independent factors of the face data. 

\begin{table}[ht!]
\caption{Average disentangling score of models with the disentangling metric proposed in \cite{factor-vae}.}
\centering
\begin{tabular}{|l|c|c|c|}
\hline
    & VAE & Factor-VAE & WVAE \\ 
\hline
\small{Disentangling score} &  80.0   & 82.0  & 85.0 \\ 
\hline
\end{tabular}
\label{table:results}
\end{table} 

Table. \ref{table:results} summarizes the quantitative analysis results on the 2D Shapes dataset from the three different deep generative models (VAE, Factor-VAE, and WVAE) with the disentanglement metric proposed in \cite{factor-vae}. The disentangling score is the accuracy of major vote classifer. Therefore, higher disentangling score means the model encodes the generative factors into the dimension of the latent variable more independently. Our proposed model's disentanglement score is higher than ones of original VAE and Factor-VAE. Note that the evaluation metric is the one proposed in the Factor-VAE paper \cite{factor-vae}, which is described in Table. \ref{table:metric}.  

\begin{figure}[ht!]
\centerline{\hbox{ 
\includegraphics[width=1.3in]{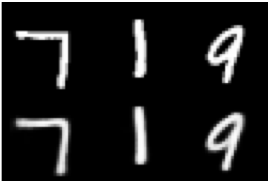}\hspace{0.1in}
\includegraphics[width=1.3in]{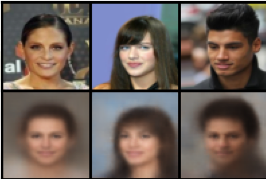}\hspace{0.1in}
\includegraphics[width=1.3in]{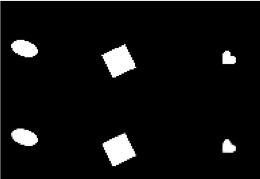}
}}
\caption{Reconstruction results of WVAE on the three datasets. In each image, the top row is original images and the bottom row is the reconstructed ones.}
\label{fig:recon}
\end{figure}

\begin{figure}[ht!]
\centerline{\hbox{ 
\includegraphics[width=3.8in]{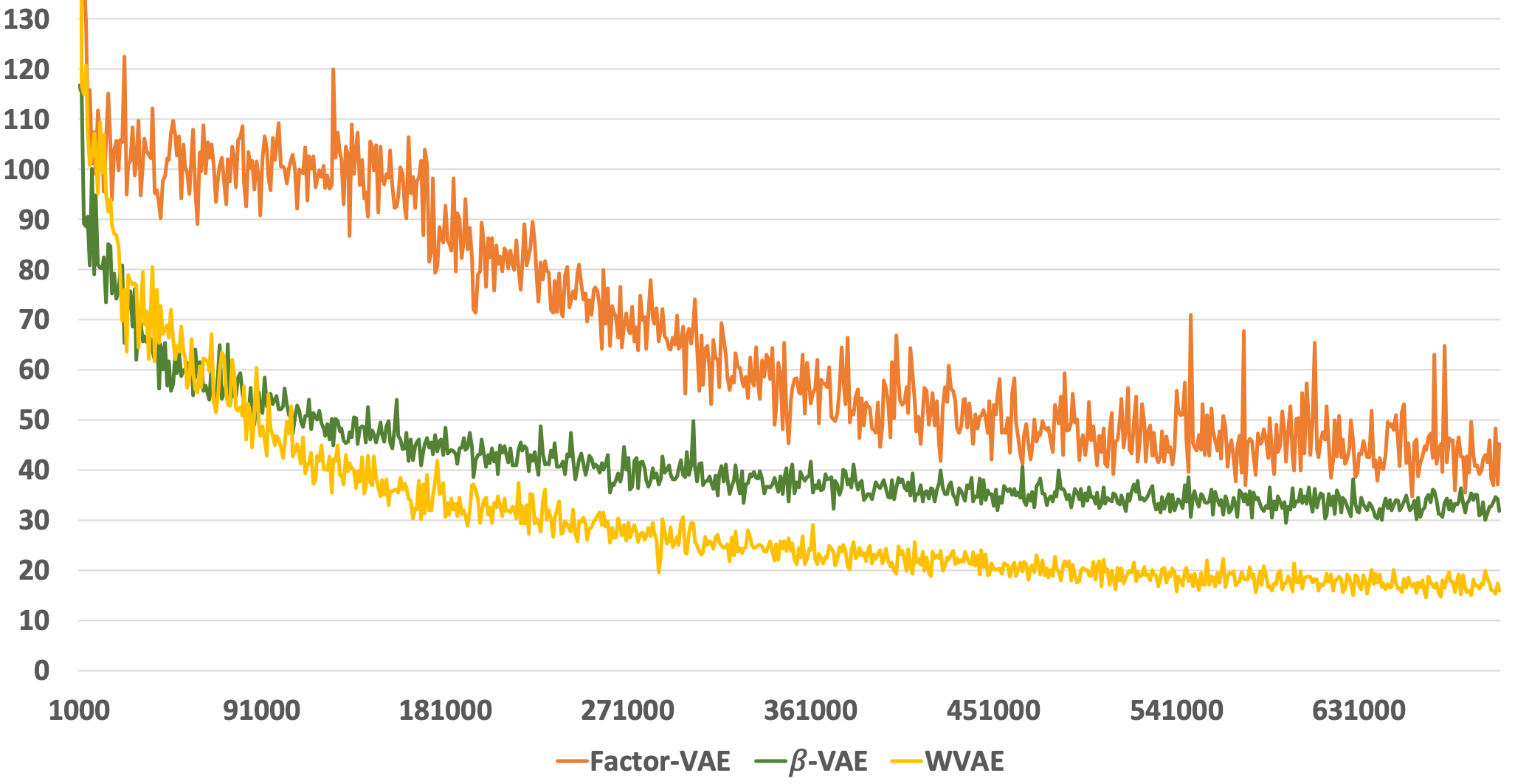}
}}
\centerline{\hbox{ 
(a) Reconstruction error of 2D Shape}}
\centerline{\hbox{ 
\includegraphics[width=3.8in]{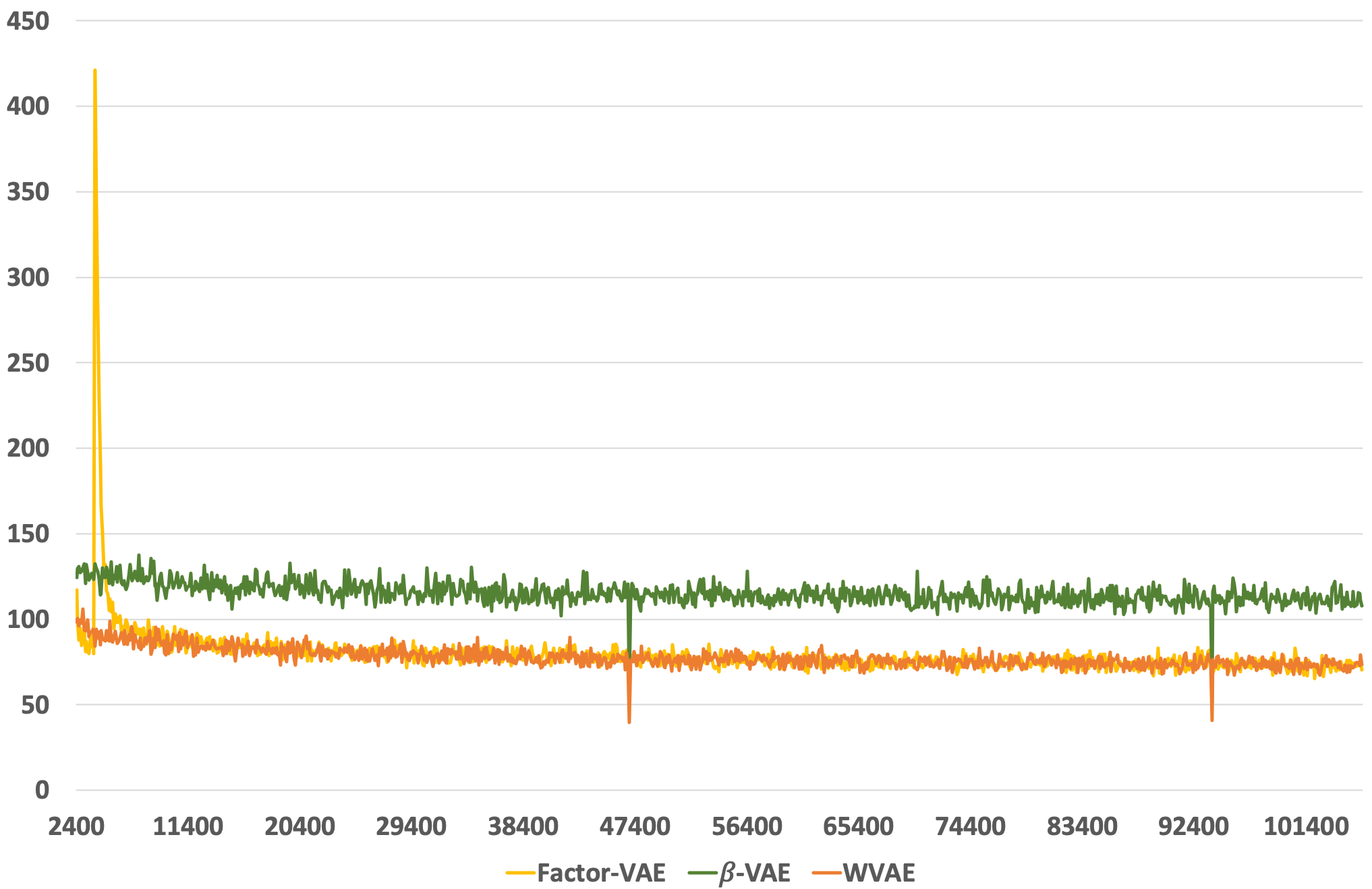}
}}
\centerline{\hbox{ 
(b) Reconstruction error of MNIST}}
\caption{Training curves of reconstruction error for the three models on the 2D Shape and MNIST dataset.}
\label{fig:recon_err}
\end{figure}

To check reconstruction quality, Fig. \ref{fig:recon} shows the reconstruction results of WVAE model for CelebA, MNIST, and 2D Shapes. As we mentioned in Section \ref{sec:whitening_latent}, WVAE model has the same reconstruction quality as original VAE. Also, we compare the reconstruction error of three models: Factor-VAE ($\gamma=6.4$), $\beta$-VAE (2D Shape: $\beta=4$, MNIST: $\beta=6$), and WVAE during training, as shown in Fig. \ref{fig:recon_err}, where our model's error is lower than the other two models'. This proves that our proposed method does not sacrifice reconstruction quality while obtaining the more disentangled latent variable.

\begin{figure}[ht!]
\centerline{\hbox{ 
\includegraphics[width=2.7in]{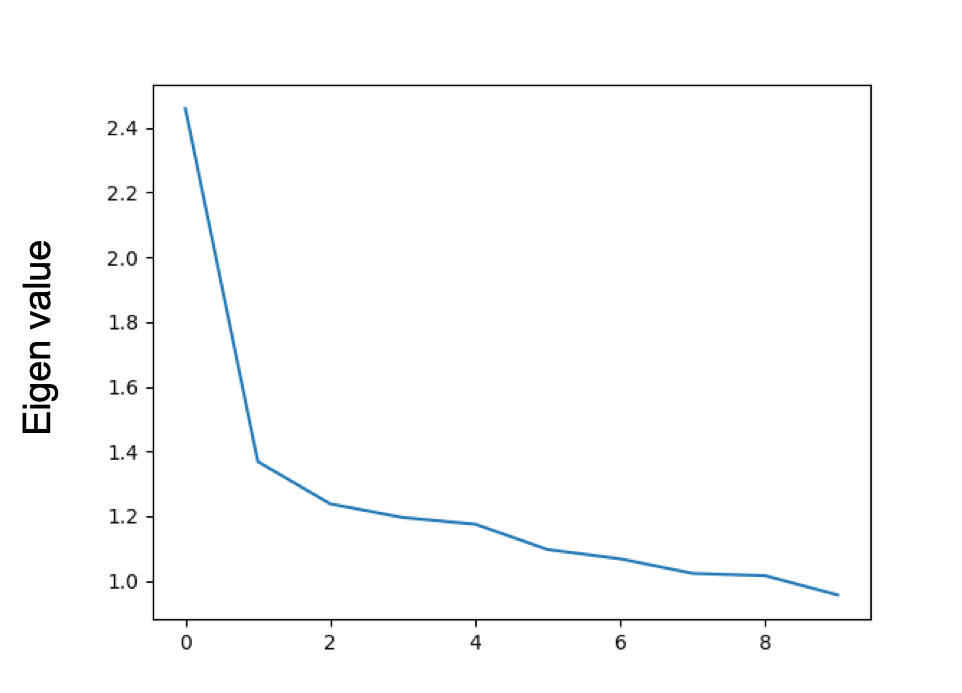}
\includegraphics[width=2.7in]{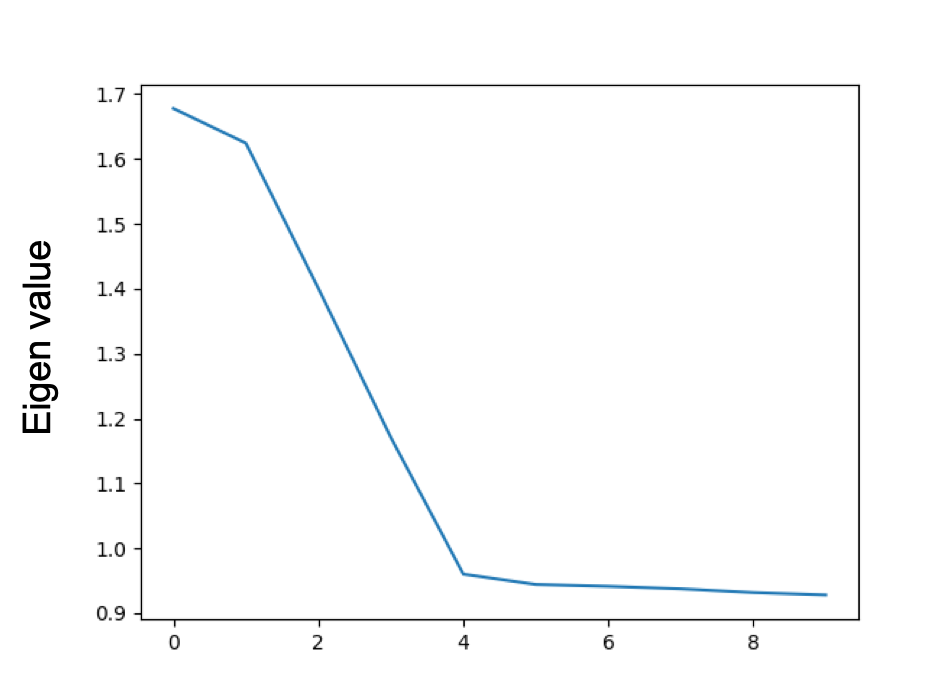}
}}
\centerline{\hbox{ 
(a) CelebA  \hspace{1.7in} (b) 2D Shapes  }}
\caption{Eigenvalue of the latent variable of the CelebA training dataset and the 2D Shapes dataset. The eigenvalues are sorted by descending order. Note that only a few factors are dominant.}
\label{fig:eigen_value}
\end{figure}

In addition to the experiment results above, using our proposed WVAE model has an advantage to interpret the meaning of latent variable and factors of generated images. To apply whitening method to the latent variable, we calculate the eigenvalue and eigenvector of the latent variable. Fig. \ref{fig:eigen_value} presents the eigenvalues of the latent space for the CelebA training dataset and the 2D Shapes dataset. It is shown that a few latent variables dominate the latent space, meaning that the factors in the latent space by VAE have strong correlation to each other. This eigenvalue analysis can indicate dominant dimensions of the latent variable and the generating factors corresponding to those dominant dimensions. Such knowledge is important to make a model more explainable.

\section{Conclusion}
\label{sec:conclusion}
Learning a disentangled representation of given data set is important for not only deep generative models but also making an artificial intelligence (AI) model to understand our real world more conceptually. We showed that the latent variables can be disentangled by a simple method (PCA whitening) without any change in the objective function or model architecture. The results of qualitative and quantitative analysis show that our proposed method can disentangle the latent variable as good as other disentangling models can, without loss of reconstruction quality. Also, with eigenvalue analysis, we could see that our proposed method makes a more interpretable model. For the future work, we will try other transformation methods like non-negative matrix factorization (NMF) or independent component analysis (ICA).

\section*{Acknowledgement}
This research was supported by Basic Science Research Program through the National Research Foundation of Korea(NRF) funded by the Ministry of Education (2017R1D1A1B03033341), and by Institute for Information \& communications Technology Promotion(IITP) grant funded by the Korea government(MSIT) (No. 2018-0-00749, Development of virtual network management technology based on artificial intelligence).

\bibliographystyle{splncs04}
\bibliography{hchoi2018.bib}

\end{document}